# Source Separation and Higher-Order Causal Analysis of MEG and EEG

Kun Zhang[1,2]
[1]Max Planck Institute for Biological Cybernetics
Spemannstr. 38, 72076 Tübingen
Germany

Aapo Hyvärinen[2,3]
[2]Dept. of Computer Science and HIIT
[3]Dept. of Mathematics and Statistics
University of Helsinki
Finland

## Abstract

Separation of the sources and analysis of their connectivity have been an important topic in EEG/MEG analysis. To solve this problem in an automatic manner, we propose a two-layer model, in which the sources are conditionally uncorrelated from each other, but not independent; the dependence is caused by the causality in their time-varying variances (envelopes). The model is identified in two steps. We first propose a new source separation technique which takes into account the autocorrelations (which may be time-varying) and time-varying variances of the sources. The causality in the envelopes is then discovered by exploiting a special kind of multivariate GARCH (generalized autoregressive conditional heteroscedasticity) model. The resulting causal diagram gives the effective connectivity between the separated sources; in our experimental results on MEG data, sources with similar functions are grouped together, with negative influences between groups, and the groups are connected via some interesting sources.

## 1 INTRODUCTION

Blind source separation (BSS) of the magnetoencephalography (MEG) or electroencephalography (EEG) data has been a very active research area (Baillet et al., 2001). As a widely-used BSS technique, independent component analysis (ICA, Hyvärinen et al., 2001) has been found very useful to find and remove artifacts (Jung et al., 2000). However, it is difficult to find components related to brain activity. This may be due to the lack of independence or non-Gaussianity of the activations of different cell assemblies involved (Hyvärinen et al., 2010). Some second-order statistics (SOS)-based BSS techniques, such as SOBI (Second-order Blind Identification, Belouchrani et al., 1997), assume that the sources are uncorrelated and have different autocorrelations (or spectra). Since some assemblies may have very similar autocorrelations, it is also very difficult for such methods to separate brain activity-related components.

In this paper, we aim to find more specific properties of the EEG/MEG sources, and to provide suitable machine learning tools for analysis. In particular, we propose a two-layer generative model for EEG/MEG signals. Its identification enables BSS and discovery of the effective connectivity between the sources implied in their envelopes. In the first layer of this model, the observable scalp sensor signals are assumed to be linear mixtures of underlying sources, which are conditionally uncorrelated from each other. Each source follows an autoregressive (AR) model, and the coefficients may be time-varying. In the second layer, statistical dependencies between the sources are introduced; their time-varying conditional variances (or envelopes) are correlated. The GARCH (generalized autoregressive conditional heteroscedastic) model (Bollerslev, 1986), in combination with the idea of causality in variance (Granger et al., 1984), gives a special kind of multivariate GARCH model; estimation of this model produces the variances of the sources and the causal relations among them.

The model is identified conveniently in two steps. The first step performs source separation, and we propose a new and unified SOS-based method, which could separate the sources if they have different AR coefficients (which may be time-varying), *or* if their time-varying conditional variances are not proportional to each other over time. In the second step, we select a subset of the estimated sources which are likely to correspond to brain activities and have time-varying variances; we then identify the causal relations among the variances of these sources. Alternatively, one can decompose the estimated envelopes of the sources into some components or "modulators" (which may be closely related to particular stimuli), and perform clustering of the sources according to the relationships between the envelopes and the estimated modulators.

Finally, we use the proposed method to analyze the MEG recordings during naturalistic stimulation. According to the resulting effective connectivity between the sources, the sources are automatically divided into two groups, and sources with similar functions are in the same group. There exist positive influences mainly inside the groups, and negative influences between the groups. Moreover, by decomposing the envelopes with the proposed source separation method, we obtain the modulators underlying the source envelopes. The decomposition results are consistent with the grouping derived from the causal diagram.

## 2 MODELLING EEG/MEG: UNCORRELATED SOURCES & CAUSALITY IN VARIANCES

### 2.1 MIXING PROCEDURE

As in ordinary BSS, we assume that the vector of the observable EEG/MEG recordings, $\mathbf{x}(t) = (x_1(t), ..., x_N(t))^T$, is a linear transformation of the vector of the underlying sources $\mathbf{s}(t) = (s_1(t), ..., s_N(t))^T$:

$$\mathbf{x}(t) = \mathbf{A}\mathbf{s}(t), \quad (1)$$

where $\mathbf{A}$ is assumed to be of full rank, such that all sources can be recovered from $\mathbf{x}(t)$. Here for simplicity, we have assumed that the sources and observed signals have the same number. (Note that usually we need to reduce the dimensionality of the raw EEG/MEG sensor signals to obtain $\mathbf{x}(t)$).

The sources usually have significant autocorrelations, which may be time-varying. Here we assume that the sources can be modelled by a AR($L$) model:[1]

$$s_i(t) = \sum_{\tau=1}^{L} c_{i\tau,t} s_i(t-\tau) + e_i(t), \quad (2)$$

where $c_{i\tau,t}$ are the (time-varying) coefficients, and innovations (or errors) $e_i(t)$ are temporally white and uncorrelated with $e_j(t), j \neq i$, and follow the Gaussian distribution with zero mean and variance $\sigma_{it}^2$. That is, the conditional distributions of $s_i(t)$ is $p(s_i(t)|s_i(t-$

[1]Here we assume that the data have been made zero-mean, and for simplicity of the presentation, we assume that the constant term in the AR model is zero. If needed, one can incorporate it in the model, without complicating the algorithm.

$k), k > 0) = \mathcal{N}(\sum_{\tau=1}^{L} c_{i\tau} s_i(t-\tau), \sigma_{it}^2)$, or equivalently,

$$e_i(t) = \sigma_{it} z_i(t), \text{ where } z_i(t) \sim \mathcal{N}(0,1). \quad (3)$$

Now let us specify the time-varying variances $\sigma_{it}^2$.

### 2.2 MOELLING SOURCE ENVELOPES BY EXTENDING GARCH MODELS

Figure 1 (top) shows the time course and autocorrelations in the squared values of a typical MEG signal after whitening, which was obtained as the innovation of the AR(10) model. One can see that the innovation, although temporally uncorrelated, has significantly positive autocorrelations in the squared values. This is a well-established phenomenon for financial return series such as stock returns, known as "volatility clustering". That is, the variance changes over time, and large (small) changes in the time series tend to be followed by large (small) changes of either sign.

The autoregressive conditional heteroscedasticity (ARCH)-type models (Engle, 1982) were proposed to model the time-varying variance as a weighted sum of the squared values of the past innovations and some constant. To estimate the variance accurately, ARCH usually requires fairly many lags of the past data. As a powerful extension of ARCH, the generalized ARCH (GARCH) model (Bollerslev, 1986) avoids this problem by further incorporating the past variances in the model. Using the GARCH($p,q$) model, we can express the conditional variance of $s_i(t)$ as follows:

$$\sigma_{it}^2 = \omega_i + \sum_{\tau=1}^{q} \alpha_{i\tau} e_i^2(t-\tau) + \sum_{\tau=1}^{p} \beta_{i\tau} \sigma_{i,t-\tau}^2, \quad (4)$$

where restrictions $\omega_i > 0$, $\alpha_{i\tau} \geq 0$, and $\beta_{i\tau} \geq 0$ are imposed to ensure $\sigma_{it}^2$ to be positive. In practice, GARCH(1,1) is usually adequate. This model has been proven very useful in modelling and forecasting the time-varying variances of financial returns. The middle row in Figure 1 shows the time-varying variances of the innovations of the MEG signal estimated by GARCH(1,1). The corresponding standardized residual, which was obtained by dividing the innovation by the time-varying standard deviation, is plotted in the bottom row. Clearly, compared to the original innovation series, it is much closer to be i.i.d., and the autocorrelations in its squared values are almost zero.

When we have parallel time series, whose variances are dependent, one series may be useful to predict the variance, or it may *cause* the change of the variance, of another one. To formulate that, by incorporating the idea of causality in variance (Granger et al., 1984), we extend GARCH to the following constrained multivariate GARCH model, which expresses the conditional

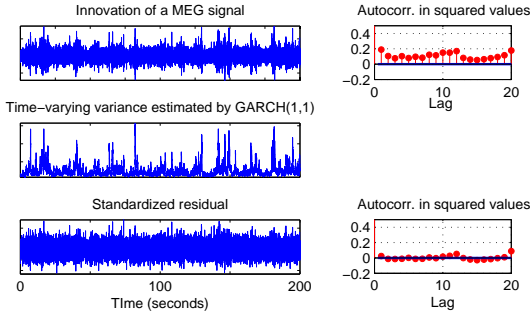

Figure 1: An illustration of estimating the envelope of the innovation of a MEG signal. From top to bottom: innovation, time-varying variance estimated by GARCH(1,1), and standardized residual. Right column: autocorrelation coefficients in squared values.

variance of the $i$th series as

$$\sigma_{it}^2 = \omega_i + \sum_{j=1}^{N}\sum_{\tau=1}^{q} \alpha_{ij,\tau} e_j^2(t-\tau) + \sum_{\tau=1}^{p} \beta_{i\tau} \sigma_{i,t-\tau}^2. \quad (5)$$

There exist many different forms of multivariate GARCH models; this one is highly constrained, tailored to do causal discovery in the envelopes. Here we term it causal-in-variance GARCH (CausalVar-GARCH). Compared to standard GARCH (Eq. 4), this model contains extra terms $\sum_{j\neq i}\sum_{\tau=1}^{q} \alpha_{ij,\tau} e_{j,t-\tau}^2$; the parameters $\alpha_{ij}$ indicate the (causal) influence from $s_{jt}$ to $s_{it}$ in the variances. If $\alpha_{ij} \neq 0$, $s_j(t)$ is said to be causal in variance to $s_i(t)$ (Granger et al., 1984). This model is closely related to Granger causality analysis (Granger, 1980). However, here we prefer not to perform ordinary Granger causality analysis on $e_i^2(t)$, since it is difficult to find a suitable functional form to represent $e_i^2(t)$ in terms of $e_j^2(t-\tau)$. Moreover, $\sigma_{it}^2$ are interesting to us and we would like to estimate them from the data. In practice, it is expected that the causal relations are sparse, i.e., only a small subset of $\{\alpha_{ij,\tau}\}$ in CausalVar-GARCH are non-zero. To achieve this, we use the model selection approach proposed by Zhang et al. (2009), which exploits the data-adaptive $\ell_1$ penalization with a fixed penalization parameter.

Finally, the combination of Eqs. 1, 2, 3, and 5 gives the two-layer generative model for the EEG/MEG signals. The first layer generates the observed signals as linear combinations of the sources, and the second layer models the conditional variance of each source. All involved parameters can be estimated simultaneously by maximum likelihood. However, the parameters at different stages (including $\mathbf{A}$, parameters in CausalVar-GARCH Eq. 5, and those in the AR model Eq. 2) interfere with each other, which may cause estimation difficulties. Therefore, we prefer estimation in two steps, as explained next.

## 2.3 ESTIMATION IN TWO STEPS

One can first perform source separation and then analyze the source envelopes. The procedure is as follows.

1. Separate the EEG/MEG sources using a suitable BSS method, which will be proposed in Section 3.

2. First, remove estimated sources which are clearly artifacts (e.g. based on visual inspection, or the methods in Hyvärinen et al., 2010). Since we aim to analyze the information hidden in the time-varying variances of the sources, sources with constant variances, if there are any, should be excluded; we use Engle's Lagrange Multiplier (LM) test (Engle, 1982) to test for the existence of ARCH behavior in the sources. Then, use the CausalVar-GARCH model to find the causality in the envelopes of the selected sources.

As a result, we find possibly interesting sources and their effective connectivity implied in their envelopes.

## 2.4 DECOMPOSITION OF ENVELOPES

Another way to analyze the dependent source envelopes is to decompose them into some uncorrelated and simple components. The envelopes can be considered as products of some underlying "modulators" with different strength. Mathematically, we assume that the modulation process can be written as

$$\sigma_{it}^2 = \prod_{k=1}^{K}(e^{v_{kt}})^{d_{ik}}, \text{ or equivalently, } \log \sigma_{it}^2 = \sum_{k=1}^{K} d_{ik} v_{kt},$$

where $e^{v_{kt}}, k = 1, ..., K$ are the underlying modulators, and $d_{ik}$ denote the strength of the influence of the $k$th modulator on the envelope $\sigma_{it}^2$. As $\log \sigma_{it}^2$ *follow a linear mixing model*, one can then decompose $\log \sigma_{it}^2$ with suitable BSS methods to find the modulators. In particular, since the modulators are strongly autocorrelated, BSS methods based on the temporal information of the data are expected to work well. In our experiments, the BSS method proposed in Section 3 is employed to do such decomposition. We expect that some modulators may be physically interpretable; for instance, they may be closely related to stimuli. One can also interpret the $d_{ik}$ as giving $K$ groupings of the sources, by finding for each $k$ those $d_{ik}$ which are large enough, and sources in the same group may have some similar functions.

## 3 A UNIFIED SOS-BASED BSS METHOD IN TIME DOMAIN

ICA (Hyvärinen et al., 2001) requires the sources to be separated independent and non-Gaussian. SOBI (Be-louchrani et al., 1997) could separate the sources if they have different autocorrelations. On the other hand, if the sources are locally uncorrelated and their local variances fluctuate somewhat independently of each other (Matsuoka et al., 1995; Pham & Cardoso, 2001), one can recover the sources, by making the outputs locally uncorrelated. These requirements or assumptions may not be satisfied by the model given in Section 2; in particular, for EEG/MEG signals, some sources may have similar autocorrelations (or frequency spectra), and the variances of some sources may be approximately constant. Some algorithms in frequency domain or time-frequency domain have also been proposed (Hosseini et al., 2009; Pham & Cardoso, 2003). Such algorithms require a good estimate of the frequency spectral densities and need to handle imaginary numbers. Here to estimate the two-layer model given in Section 2, we propose a unified SOS-based time-domain BSS method, which could separate conditionally uncorrelated sources if the local variances of the sources change somewhat independently (as in Matsuoka et al., 1995), *or* if the sources have different autocorrelations (which are allowed to be time-varying). The method is derived by maximum likelihood (Pham & Cardoso, 2001), so generally speaking, the estimate is statistically appealing.

### 3.1 METHOD

The model for source separation considered here is the combination of Eqs. 1, 2, and 3; here we do not assume the GARCH-type model for each source, but simply assume that some source innovations have time-varying variances $\sigma_{it}^2$. We aim to recover the sources $s_i(t)$ in Eq. 1 using $\mathbf{y}(t)$, which is obtained by applying the linear transformation $\mathbf{W}$ on $\mathbf{x}(t)$:

$$\mathbf{y}(t) = \mathbf{W}\mathbf{x}(t). \qquad (6)$$

As usual, we assume that the variances $\sigma_{it}^2$ and the AR coefficients $c_{i\tau,t}$ change smoothly over time, such that they could be approximately considered as constants in a short window.

Suppose we divide all time points at $t = 1, ..., T$ into $M$ blocks $\mathcal{T}_1, ..., \mathcal{T}_M$. Denote by $K$ the length of each block. Denote by $\sigma_i(m)^2$ and $c_{i\tau}(m)$ the innovation variances $\sigma_{it}^2$ and the coefficients $c_{i\tau,t}$ in the $m$th block, respectively.

The conditional distributions of $s_i(t)$ are $p(s_i(t)|s_i(t-k), k > 0) = \mathcal{N}(\sum_{\tau=1}^L c_{i\tau} s_i(t-\tau), \sigma_{it}^2)$, and the conditional distribution of $\mathbf{x}(t)$ is $p(\mathbf{x}(t)|\mathbf{x}(t-k)) = p(\mathbf{s}(t)|\mathbf{s}(t-k))/|\mathbf{A}| = \prod_{i=1}^n p(s_i(t)|s_i(t-k))/|\mathbf{A}|$. After simplifications, the negative data (conditional) likelihood becomes $-\log p(\mathbf{x}(1), ..., \mathbf{x}(T)) = \frac{1}{2}\sum_{i=1}^n \sum_{l=1}^m \sum_{t\in\mathcal{T}_m}\Big[n\log(2\pi) + \log\sigma_i(m)^2 +$ $\tilde{y}_i(t)^2/\sigma_i(m)^2\Big] - T\log|\mathbf{W}|$, where $\tilde{y}_i(t)$ denote the innovations of fitting $y_i(t)$ with the AR model Eq. 2, i.e., $\tilde{y}_i(t) = y_i(t) - \sum_{\tau=1}^L c_{i\tau}(m)y_i(t-\tau)$. Like in Pham and Cardoso (2003), by minimizing the negative likelihood w.r.t. $\sigma_i(m)^2$, one can find its estimate: $\widehat{\sigma_i(m)^2} = \sum_{t\in\mathcal{T}_m}\tilde{y}_i(t)^2/K$. Substituting it back into the negative likelihood leads to (with certain constants dropped)

$$J = \frac{K}{2}\sum_{i=1}^n\sum_{m=1}^M \log\widehat{\sigma_i(m)^2} - T\log|\mathbf{W}|. \qquad (7)$$

The autocovariance matrices of $\mathbf{x}(t)$ in each block will be involved in our algorithm. To make them symmetrical, we define the autocovariance matrice of $\mathbf{x}_t$ in the $m$th block at lag $d$ as $\mathbf{R}_{\mathbf{x},d}^{(m)} \triangleq \frac{1}{2K}\sum_{t\in\mathcal{T}_m}\big[\mathbf{x}(t)\mathbf{x}(t+d)^T + \mathbf{x}(t+d)\mathbf{x}(t)^T\big]$. Let $\gamma_{y_i,d}^{(m)}$ be the $d$th-order autocovariance of $y_i(t)$ in the $m$th block, i.e., $\gamma_{y_i,d}^{(m)} = \sum_{t\in\mathcal{T}_m} y_i(t)y_i(t+d)/K$. They can be directly calculated from $\mathbf{R}_{\mathbf{x},d}^{(m)}$ as $\gamma_{y_i,d}^{(m)} = \mathbf{w}_i^T \mathbf{R}_{\mathbf{x},d}^{(m)} \mathbf{w}_i$, where $\mathbf{w}_i^T$ denotes the $i$th row of $\mathbf{W}$. The matrix $\mathbf{K}_i^{(m)}$ defined below consists of $\gamma_{y_i,d}^{(m)}$ as its entries:

$$\mathbf{K}_i^{(m)} \triangleq \begin{pmatrix} \gamma_{y_i,0}^{(m)} & \gamma_{y_i,1}^{(m)} & \cdots & \gamma_{y_i,L-1}^{(m)} \\ \gamma_{y_i,1}^{(m)} & \gamma_{y_i,0}^{(m)} & \cdots & \gamma_{y_i,L-2}^{(m)} \\ \vdots & \ddots & \ddots & \vdots \\ \gamma_{y_i,L-1}^{(m)} & \cdots & \gamma_{y_i,1}^{(m)} & \gamma_{y_i,0}^{(m)} \end{pmatrix}. \qquad (8)$$

After tedious calculations, we can find the derivative of Eq. 7 w.r.t. $\mathbf{w}_i$:

$$\frac{1}{T}\frac{\partial J}{\partial \mathbf{w}_i} = \frac{1}{M}\sum_{m=1}^M \frac{1}{\widehat{\sigma_i(m)^2}}\Big(\sum_{\tau_1=0}^L \sum_{\tau_2=0}^L \hat{c}_{i\tau_1}(m)$$
$$\cdot \hat{c}_{i\tau_2}(m) \cdot \mathbf{R}_{\mathbf{x},|\tau_1-\tau_2|}^{(m)}\Big)\mathbf{w}_i - [\mathbf{W}^{-1}]_{\cdot i}, \quad (9)$$

where $\hat{c}_{i0}(m) \triangleq -1$, $(\hat{c}_{i1}(m), ..., \hat{c}_{iL}(m))^T = [\mathbf{K}_i^{(m)}]^{-1} \cdot (\gamma_{y_i,1}^{(m)}, ..., \gamma_{y_i,L}^{(m)})^T$, $\widehat{\sigma_i(m)^2} = \gamma_{y_i,0}^{(m)} - (\gamma_{y_i,1}^{(m)}, ..., \gamma_{y_i,L}^{(m)}) \cdot \hat{\mathbf{c}}_i(m)$, and $[\mathbf{W}^{-1}]_{\cdot i}$ denotes the $i$th column of $\mathbf{W}^{-1}$. The corresponding natural gradient can be obtained by multiplying the right-hand side of the gradient $\frac{\partial J}{\partial \mathbf{W}}$ with $\mathbf{W}^T\mathbf{W}$ (Cichocki & Amari, 2003). As input, the algorithm just requires the local autocovariances $\mathbf{R}_{\mathbf{x},|\tau_1-\tau_2|}^{(m)}$; it is then termed as "L-ACOV I" (local autocovariance-based method I).

### 3.2 SPECIAL CASE WITH CONSTANT AR COEFFICIENTS

In some situations, $c_{i\tau,t}$ are approximately constant along time. Moreover, if the sample size is not large enough, the model with time-varying $c_{i\tau,t}$ has too much freedom, causing the danger of overfitting. It would then be better to constrain them to be constant.

In this case, AR coefficients $c_{i\tau}(m)$ become $c_{i\tau}$, which can be learned together with **W**, by minimizing Eq. 7, using an alternating optimization technique. In each iteration, we first fix $\hat{c}_{i\tau}$, and update **W** using the gradient (or natural gradient)-based method. The gradient of Eq. 7 w.r.t. $\mathbf{w}_i$ is actually similar to Eq. 9. After updating **W**, we fix **W** and update $\hat{c}_{i\tau}$, the estimate of the AR parameters for $y_i(t)$. The derivative of Eq. 7 w.r.t. $\mathbf{c}_i = (c_{i1},...,c_{iL})^T$ can be easily found. One can then update $\hat{\mathbf{c}}_i$ with the gradient-based method, or even in closed form. Details are skipped. This algorithm is termed as "L-ACOV II".

In fact, one can easily incorporate different prior knowledge or constraints in the model discussed in Section 3.1 to simplify the algorithm, or to improve the separation performance. In particular, if the innovation variance $\sigma_{it}^2$ of each source is constant, the nonstationarity of the autocorrelations in the sources could also enable source separation.

### 3.3 SIMULATIONS

To make the simulations of practical use for EEG/MEG analysis, we generated the data whose properties are similar to real EEG/MEG data. We took into account various effects, including non-stationarity, source autocorrelations (which may be time-varying), correlation in the envelopes, and the noise effect. In Simulation 1, 10 sources were first generated as independent AR(4) processes with Gaussian errors, and the coefficients in the AR model were randomly chosen in the range $[-0.1, 0.2]$. Next, four sources were further modulated by the same modulator with different strengths; the modulator was a sinusoid waveform, and a positive number was added to make it positive. They were then mixed by a randomly generated mixing matrix **A**. Finally, uncorrelated Gaussian noise with the signal-to-noise ratio 30dB was added to the mixtures. The sample size was 4000. The methods for comparison were SOBI (Belouchrani et al., 1997), FastICA (Hyvärinen, 1999) with the tanh nonlinearity and in the symmetrical manner, JADE (Cardoso & Souloumiac, 1993), the method by joint diagonalization of local covariances (Hereafter denoted by 'JD-COV'), which is based on the assumption of local uncorrelatedness and nonstationary variances of the sources (Pham & Cardoso, 2001), the method exploiting the time-frequency diversity of the sources (denoted by 'TF', Pham & Cardoso, 2003), L-ACOV I (Subsection 3.1), and L-ACOV II (Subsection 3.2). In the last four methods, the data were divided into 20 blocks. For TF, discrete Fourier transform (DFT) was used to estimate the spectral density and the frequency plane was divided into 6 equi-spaced intervals.

The performance was evaluated by the Amari performance index $P_{err}$ (Cichocki & Amari, 2003), which measures how far **WA** is from a generalized permutation matrix. The smaller $P_{err}$, the better the separation performance. We repeated the simulations for 40 replications, in each of them the sources and the mixing matrix were randomly chosen. Fig. 2 (top) gives the boxplot of the Amari performance index for each method. One can see that SOBI is better then FastICA and JADE, but the remaining four are better than SOBI. JD-COV solely makes use of the time-varying local variances of the sources and neglects the autocorrelations, and its performance is not as good as those combining the time-varying conditional variances and autocorrelations of the sources. L-ACOV I and L-ACOV II are among the best. Due to the prior knowledge of the constant AR coefficients, L-ACOV II is slightly better then L-ACOV I.

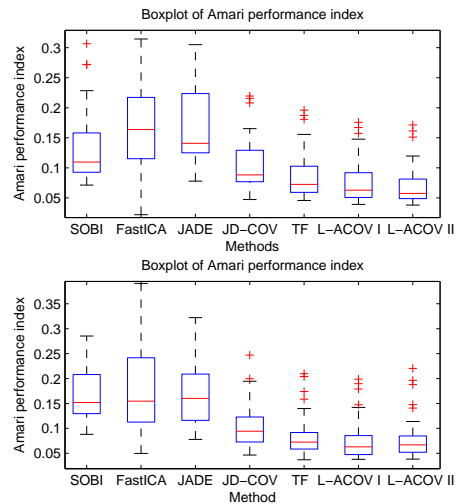

Figure 2: Boxplots of the Amari performance index in simulations. Top: Simulation 1. Bottom: Simulation 2.

The settings in the second simulation were similar to in the first one; however, the coefficients in the AR model for each source were not constant. We divided the whole time period into two parts, and for each source, in the second part we re-generated the AR coefficients randomly. The performance of different methods is shown in Fig. 2(bottom). The performance of SOBI and L-ACOV II is not as good as in Simulation 1, due to the changing autocorrelations of the sources. L-ACOV I, L-ACOV II, and TF are among the best.

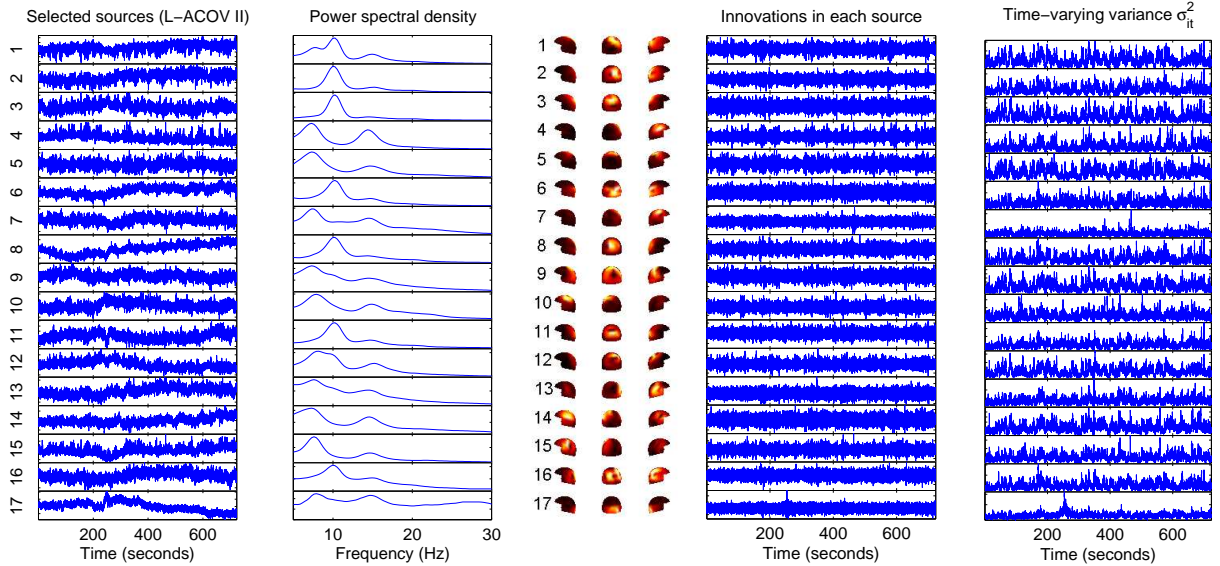

Figure 3: Selected sources estimated by L-ACOV II. Panels from left to right: waveforms (without pre-whitening), power spectral densities, topographic helmet plots, innovations, and estimated envelopes of the sources.

## 4  EXPERIMENTAL RESULTS

### 4.1  DATA AND PREPROCESSING

We applied our model and method on real MEG data recorded at the Brain Research Unit of the Helsinki University of Technology.[2] The raw recordings consisted of 204 gradiometer channels and last about 12 minutes, obtained from a healthy volunteer, who received non-overlapping auditory, visual, and tactile stimuli (Malinen et al., 2007). The data were down-sampled to 75Hz. Since some recordings have clear trends, as preprocessing, we concatenated all MEG signals and fitted an AR(10) model on the concatenated data. We then worked on the errors. This is equivalent to apply a whitening filter, which is the same for all channels, on the raw recordings. Since the recordings of different channels usually have different autocorrelations, each channel is not necessarily white. But the trends in the data disappeared and the autocorrelations common in all channels were eliminated. We then used principal component analysis (PCA) to reduce the dimensionality to $N = 40$.

### 4.2  SOURCE SEPARATION

We separated sources with different source separation methods, including L-ACOV I (Subsection 3.1), L-ACOV II (Subsection 3.2), JD-COV (Pham & Cardoso, 2001), SOBI, FastICA, and JADE. The last three methods have been widely used for BSS of the EEG/MEG signals.

When using L-ACOV I and II, we set the order of the AR model of each source to $L = 10$. For L-ACOV II, we divided the whole period into 150 segments; each segment then corresponds to 4.8 seconds. For L-ACOV I, in which each segment has its own AR model, to limit the model complexity and avoid overfitting, we divided the data into 120 segments. We repeated these two algorithms for 10 replications with random initializations for $\mathbf{W}$, and each algorithm always converged to almost the same solution. Furthermore, their results are similar to each other. We then mainly report the results by L-ACOV II. We selected 17 sources which are likely to correspond to brain activities, out of all 40 output components, by visual inspection on their topographical helmet plots, waveforms, and also the frequency spectra. These sources are reliable also in the sense that there were all found very accurately in all of the 10 replications. The LM test (Engle, 1982) showed that all of these selected sources have the ARCH effect at significance level 0.01, i.e., they have time-varying conditional variances. The first four columns of Figure 3 give the time courses, power spectral densities, topographical helmet plots, and innovations $e_i(t)$ of the selected sources. They were sorted according to their contributed variances, indicated by the norms of the corresponding columns of the mixing matrix.

To save space, the results by other methods are not shown. By inspection on the topographical helmet plots, we found that generally speaking, the sources produced by L-ACOV I, as well as those by L-ACOV II, have sharper locations than those produced by other methods, and are preferable.

---

[2] We are very grateful to Pavan Ramkumar for providing the data.

### 4.3 CAUSAL DISCOVERY IN THE ENVELOPES

We then estimated the CausalVar-GARCH(1,1) model (Eq. 5) for the innovations of the selected sources.[3] We used the adaptive $\ell_1$ penalty-based BIC-like model selection (Zhang et al., 2009) to eliminate insignificant causal connections implied by $\alpha_{i,j}$. All involved parameters were estimated by penalized maximum likelihood. The estimated envelopes of the sources are shown in Figure 3 (rightmost column).

The resulting parameters $\alpha_{ij}$ are shown in Figure 4. Correspondingly, the causal diagram among the source envelopes, or the effective connectivity, implied by $\alpha_{ij}$ in the CausalVar-GARCH model, is shown in Figure 5. One can see that all sources are divided into two groups by the green line (which was inserted manually). The causal influences inside each group are positive, while those between the groups are mainly negative. Sources #9 and #12 are on the boundary and their connections to both groups are positive. Some sources, such as #2, #3, #6, #8, #11, and #16, have strong interconnections.

In fact, the sources in the first group (left in the diagram) are mainly occipital and parietal components, related to visual processing and possibly spatial attention. In contrast, the sources in the second group are Rolandic, related to somatosensory and motor processing. Interestingly the sources #9 and #12 are located between these two brain areas both in the diagram and on the topographic helmet plots, and possibly interface the two main groups.

### 4.4 DECOMPOSITION AND CLUSTERING OF ENVELOPES

Next, we applied L-ACOV I (with the data divided into 100 blocks and $L = 10$) on the logarithms of the source envelopes, as suggested in Subsection 2.4, to decompose them. This produced the components $v_{kt}$ associated with the modulators and the mixing matrix $\mathbf{A}_\sigma$. The sign of each component was adjusted to make its total contribution to all $\log \sigma_{it}^2$ positive. They were sorted according to the contributed variances. Figure 6 shows some columns of the mixing matrix $\mathbf{A}_\sigma$ (the time courses of the output components $v_{kt}$ are not given).

---
[3]We also tried CausalVar-GARCH(2,3) and found that $\alpha_{ij,\tau}$ and $\beta_{i\tau}$ (with $\tau > 1$) are not significant, so we report the result with CausalVar-GARCH(1,1).

The dependencies between stimuli and the sources are very important to understand brain activation related to natural stimuli. we found that $v_{1t}$ and $v_{2t}$ are clearly correlated with the stimuli. One might further use regression or other techniques to find the combinations of the modulators which are most informative to distinguish stimulus states; this is out of the scope of this paper. From the mixing matrix (Figure 6), we can see that all envelopes, expect that of Source #17, are strongly positively related to $v_{1t}$. This is consistent with the causal diagram in Figure 5, in which Source #17 is almost isolated; Source #17 is possibly an artifact. The second column of $\mathbf{A}_\sigma$ indicates that Sources #2, #3, #6, #8, #11, which have strongly negative relations to $v_{2t}$, should be grouped together; the fifth column implies that Sources #14, #15, #4, and #7 are closely related. These are again consistent with the effective connectivity. Moreover, the sixth column shows a significant similarity between Sources #7 and #10. According to the fourth column of $\mathbf{A}_\sigma$, Sources #4 and #15 should be grouped together. In fact, Sources #7 and #10 have very similar spatial locations, but on opposite hemispheres. The same applies to Sources #4 and #15.

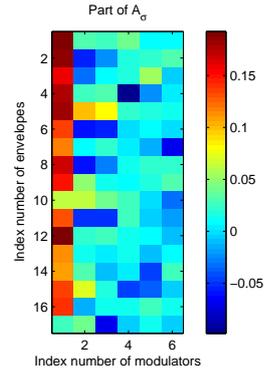

Figure 6: The first six columns of the mixing matrix $\mathbf{A}_\sigma$ obtained by decomposing $\log \sigma_{it}^2$.

## 5 CONCLUSIONS

We proposed a two-layer model as a possible way to explain the generating process of the EEG/MEG signals. Compared to other blind source separation models, this model has an additional layer to account for the dependencies in the envelopes of the sources. Causal-in-variance GARCH, as a constrained multivariate GARCH model, was employed to represent the causality in the source envelopes. Estimation of the two-layer model enables automatic source separation and causal discovery in their envelopes. To verify the causal connections, we also gave a scheme to decompose the envelopes into modulators which may be physically interpretable and help to do clustering of the sources. Experimental results on real MEG signals show that the envelopes of the separated sources are dependent, which challenges the independence assumption underlying the application of ICA to separate brain activities. The resulting effective connectivity implied in the source envelopes reveals how the sources are related to each other, and provides com-pletely new kind of information about the (causal) interactions between different brain areas.

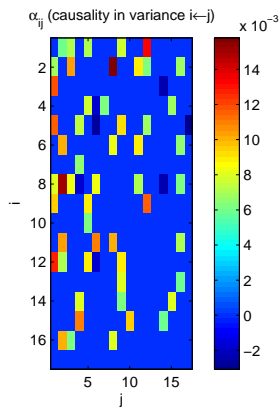
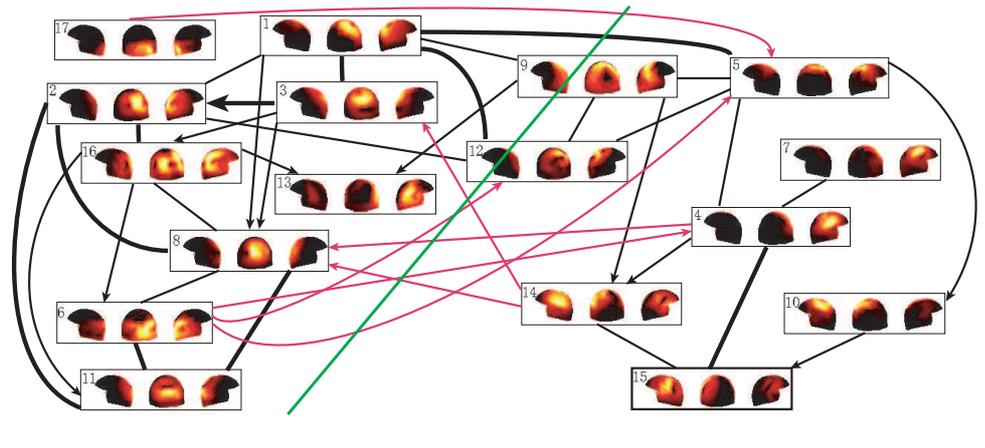

Figure 4: Estimated values of $\alpha_{ij}$ in the CausalVar-GARCH model.

Figure 5: Causality in the variances of the sources implied by $\alpha_{ij}$ in the CausalVar-GARCH model. The thickness of the lines indicates the strength of the causal effects. Undirected lines mean bi-directed causal relations. Black (red) lines show positive (negative) effects. Clearly the green line (inserted manually) divides all sources into two groups.